\documentclass[runningheads]{llncs}
\usepackage{graphicx}
\usepackage{booktabs}
\usepackage{array}
\usepackage{multirow}
\usepackage{float}
\usepackage{amsmath}
\usepackage{cleveref}
\usepackage{xcolor,rotating}
\newcommand{\datasets}{\textsc{mWikiBias} and \textsc{mWNC}}

\begin{document}
\title{Multilingual Bias Detection and Mitigation for Indian Languages}

\author{Ankita Maity\inst{1} \and
Anubhav Sharma\inst{1} \and
Rudra Dhar\inst{1} \and
Tushar Abhishek\inst{1,2} \and \\
Manish Gupta\inst{1,2} \and
Vasudeva Varma\inst{1}}
\authorrunning{A. Maity et al.}

\institute{IIIT Hyderabad, India \and Microsoft, India\\}

\maketitle
\begin{abstract}
Lack of diverse perspectives causes neutrality bias in Wikipedia content leading to millions of worldwide readers getting exposed by potentially inaccurate information. Hence, neutrality bias detection and mitigation is a critical problem. Although previous studies have proposed effective solutions for English, no work exists for Indian languages. First, we contribute two large datasets, \datasets{}, covering 8 languages, for the bias detection and mitigation tasks respectively. Next, we investigate the effectiveness of popular multilingual Transformer-based models for the two tasks by modeling detection as a binary classification problem and mitigation as a style transfer problem. We make the code and data publicly available\footnote{\url{https://tinyurl.com/debias23}\label{datafootnote}}.

\end{abstract}

\section{Introduction}


Wikipedia has three core content policies: Neutral Point of View (NPOV), No Original Research, and Verifiability\footnote{\url{https://en.wikipedia.org/wiki/Wikipedia:Core\_content\_policies}}. NPOV means that content should represent fairly, proportionately, and, as far as possible, without editorial bias, all the significant views that have been published by reliable sources on a topic\footnote{\url{https://en.wikipedia.org/wiki/Wikipedia:Neutral\_point\_of\_view}}. This means (1) Opinions should not be stated as facts and vice versa. (2) Seriously contested assertions should not be stated as facts. (3) Nonjudgmental language should be preferred. (4) Relative prominence of opposing views should be indicated. This is definition of bias we follow in this paper.

\begin{figure*}[!t]
    \centering
    \includegraphics[width=\linewidth]{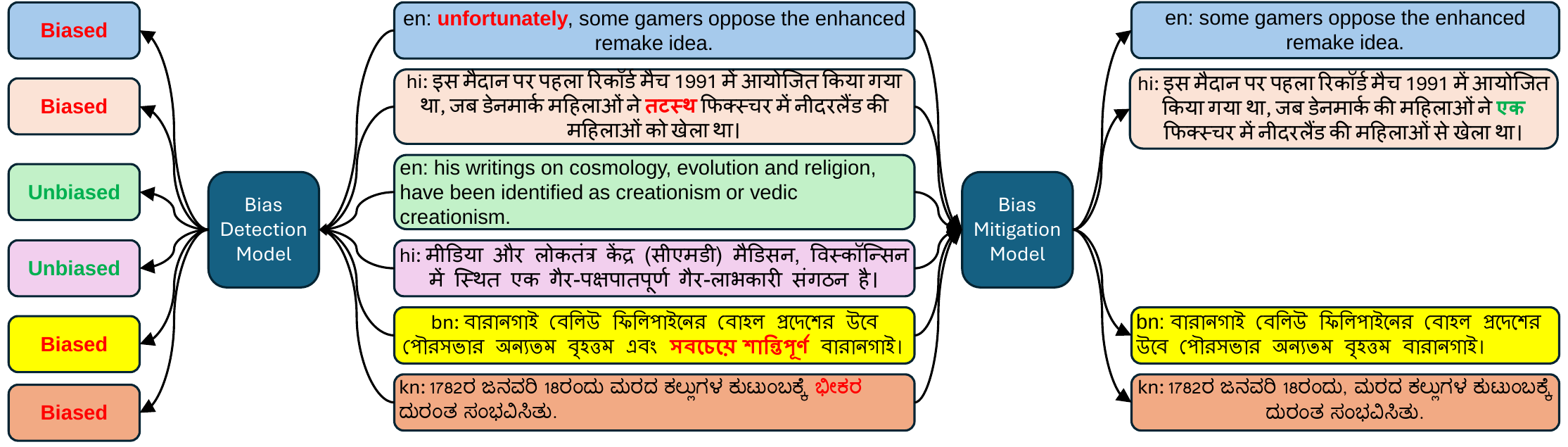}
    \caption{Bias Detection and Mitigation Examples from \textsc{mWikiBias} Dataset}
    \label{fig:examples}
\end{figure*}

Considering Wikipedia's (1) volume and diversity of content, (2) frequent updates, and (3) large and diverse userbase, automatic bias detection and suggestion of neutral alternatives is important. Bias can lead to inaccurate information or dilution of information. Particularly, lower article quality and fewer editors of Indian language Wikipedia pages makes such a system indispensable. Hence, in this work, we study how to detect sentences that violate the NPOV guidelines and convert them to more neutral sentences for Indian languages, as shown in Fig.~\ref{fig:examples}.

While there exists bias detection and mitigation studies~\cite{zhong2021wikibias,pryzant2020automatically,lai2022multilingual,liu2021mitigating} for English, there is hardly any such work for other languages. Aleksandrova et al. \cite{aleksandrova2019multilingual} work on bias detection for Bulgarian and French, but their method requires a collection of language-specific NPOV tags; and is therefore difficult to extend to Indian languages. Lastly, there exists no dataset for multilingual bias mitigation. We fill the gap in this paper by proposing two new multilingual bias detection and mitigation datasets, \datasets{}, each covering 8 languages: English (en) and seven Indian languages - Hindi (hi), Marathi (mr), Bengali (bn), Gujarati (gu), Tamil (ta), Telugu (te) and Kannada (kn).



Bias detection is challenging because certain words lead to bias if they are written in some contexts, while not in other contexts. For bias detection, we perform binary classification using MuRIL~\cite{khanuja2021muril}, InfoXLM~\cite{chi2021infoxlm} and mDeBERTa~\cite{he2022debertav3} in zero-shot, monolingual and multilingual settings. Bias mitigation is challenging because of subjectivity and context-dependence, and the models need to strike a good balance between fairness and content preservation. For bias mitigation, we perform style transfer using IndicBART~\cite{dabre2022indicbart}, mT0~\cite{muennighoff-etal-2023-crosslingual} and mT5~\cite{xue2021mt5}. These models provide strong baseline results for the novel multilingual tasks.

Overall, we make the following contributions in this paper.
\begin{itemize}
\item We propose multilingual bias detection and mitigation for Indian languages.
\item We contribute two novel datasets, \datasets{}, to multilingual natural language generation (NLG) community. Across 8 languages, they contain $\sim$568K and $\sim$78K samples for bias detection and mitigation resp.
\item Extensive experiments show that mDeBERTa outperforms MuRIL and InfoXLM for the bias detection task. On the other hand, mT5 and mT0 perform the best for bias mitigation on \datasets{} respectively. 
\end{itemize}

\section{Related Work}

Several kinds of societal biases have been studied in the literature as part of responsible AI model building~\cite{sheng2021societal}: promotional tone \cite{de2022leveraging}, puffery \cite{bertsch2021detection}, political bias \cite{fan2019plain}, and gender and racial bias \cite{field2022controlled}. In this work, we focus on a more general form of bias called as neutrality bias. Earlier work on neutrality bias detection leveraged basic linguistic features~\cite{recasens2013linguistic,hube2018detecting} while recent work uses Transformer based models~\cite{pryzant2020automatically,zhong2021wikibias}. Unfortunately, these studies~\cite{recasens2013linguistic,pryzant2020automatically,zhong2021wikibias,hube2018detecting} focus on English only. Aleksandrova et al. \cite{aleksandrova2019multilingual} work on bias detection for Bulgarian and French, but their method requires a collection of language-specific NPOV tags, making it difficult to extend to Indian languages. 

Bias mitigation is under-studied even for English~\cite{liu2021mitigating}. We contribute datasets and strong initial baseline methods towards multilingual bias mitigation.

\section{\datasets{} Datasets}

Popular bias detection and mitigation corpora in English like Wiki Neutrality Corpus (\textsc{WNC}) \cite{pryzant2020automatically} and \textsc{WikiBias} \cite{zhong2021wikibias} were created by looking for NPOV-related tags in the edit history of the English Wikipedia dumps. Both datasets have parallel sentence structures (biased sentence linked with an unbiased version). Replication of their data curation pipeline for Indian languages did not work due to a lack of frequency and consistency in tag usage for edits in the revision history of corresponding Wikipedia pages.

Hence, we translated these datasets using IndicTrans \cite{ramesh2022samanantar} to create \textsc{mWNC} and \textsc{mWikiBias} datasets for eight languages. To create cleaner datasets, we used the following heuristics to filter samples. (1) A biased and its corresponding unbiased sentence in English typically differ by very few words. Hence, we removed samples where translation of biased sentence and unbiased sentence were exactly the same for at least one of our target languages. (2) To reduce impact of translation errors, we removed samples where sentences contained regex matches for URLs, phone numbers, and email IDs. 



For every parallel translated pair of (biased, unbiased) sentence in each language $l$, we create one sample for bias mitigation dataset, and two samples (biased and unbiased) for bias detection dataset. Overall, the total number of samples for classification are 287.6K and 280.0K for \datasets{} respectively. To reduce training compute, we took a random sample from the overall bias mitigation data, leading to 39.4K and 39.0K paired samples in the \datasets{} respectively\footnote{Experiments with full bias mitigation dataset showed similar results.}. The number of samples for each language in both the datasets is consistent. Further, both of our bias detection datasets contain an equal number of biased and unbiased samples. We divide the datasets into a train/val/test split of 90/5/5.


\section{Multilingual Bias Detection and Mitigation}
We train multilingual bias detection and mitigation models using train part of \datasets{} respectively.

\noindent\textbf{Multilingual Bias Detection Method}: 
For bias detection, we finetune three Transformer encoder-only multilingual models: InfoXLM \cite{chi2021infoxlm}, MuRIL \cite{khanuja2021muril}, and mDeBERTa \cite{he2022debertav3},  with a twin linear layer setup to detect whether a sentence is biased. We experiment with three different training setups: (1) zero-shot (training only on English and testing on the other languages), (2) monolingual (one language at a time) and (3) multilingual (trained on all languages together). For fair comparisons, we use 12 layer models with a dimensionality of 768.

\noindent\textbf{Multilingual Bias Mitigation Method}: 
For bias mitigation, we finetune three multilingual encoder-decoder transformer-based models: mT5~\cite{xue2021mt5}, IndicBART~\cite{dabre2022indicbart}, and mT0~\cite{muennighoff-etal-2023-crosslingual} over the parallel corpora to perform debiasing. For fair comparisons, we use the small version of all three models for our experiments.

\noindent\textbf{Metrics}: We evaluate bias detection models using four popular binary classification metrics: accuracy (Acc), macro-precision (P), macro-recall (R) and macro-F1.

Effectiveness of bias mitigation models should be evaluated broadly on two aspects: match with groundtruth and debiasing accuracy. For measuring match with groundtruth unbiased sentences, we use standard NLG metrics like BLEU~\cite{papineni2002bleu}, METEOR~\cite{banerjee2005meteor}, chrF~\cite{popovic2015chrf} and BERT-Score~\cite{zhang2019bertscore}. We measure debiasing accuracy using ``Normalized Accuracy (NAcc)'' defined as follows. Let $N$ be the percent of ground truth sentences in the test set that are classified as ``unbiased'' by our best bias detection model. First, given a bias mitigation model, we compute the percent of its generated outputs that are classified as ``unbiased'' by our best bias detection model. Second, we normalize this quantity by $N$ and call the ratio as Normalized Accuracy (NAcc). 

A model can easily obtain high match with groundtruth by simply copying words from the input. Similarly, a model can easily obtain high NAcc score by predicting a constant highly unbiased sentence independent of the input. A good model should be able to strike a favourable tradeoff between the two aspects. Among the four metrics for computing the match, BERT-Score has been shown to be the most reliable in NLG literature, because it captures semantic match rather than just a syntactic match. Hence, we compute the harmonic mean of BERT-Score and NAcc Score and report it as HM.



\noindent\textbf{Implementation Details}: 
For MuRIL and InfoXLM, we use a learning rate of 1e-6, weight decay of 0.001, and dropout of 0.1. We trained for 15 epochs using a batch size of 320 and mixed precision training. For mDeBERTa, we use a learning rate 2e-5 with a weight decay of 0.01, keeping the other parameters the same.
We use a batch size of 12 for the bias mitigation experiments and train for 10 epochs, using early stopping with a patience of 3. We use Adafactor optimizer with a learning rate of 1e-3 for mT5 and mT0 and AdamW optimiser with a learning rate of 1e-4 for IndicBART. All models use a weight decay of 0.01.
All models were trained on a machine with 4 NVIDIA V100 GPUs having 32GB of RAM.
\begin{table*}[!t]
    \centering
    \scriptsize
    \caption{Bias Detection Results.}
\label{tab:classifier-baselines-mWIKIBIAS-mWNC}
    \begin{tabular}{|l|l|c|c|c|c|c|c|c|c|c|c|c|c|}
    \hline
\multicolumn{2}{|c|}{}&\multicolumn{4}{c|}{MuRIL}&\multicolumn{4}{c|}{InfoXLM}&\multicolumn{4}{c|}{mDeBERTa}\\
\cline{3-14}
\multicolumn{2}{|c|}{}&Acc&P&R&F1&Acc&P&R&F1&Acc&P&R&F1\\
\hline
\hline
\multirow{3.5}{*}{\rotatebox{90}{\parbox{0.75cm}{\textsc{mWiki Bias}}}}&ZeroShot&59.26&61.53&59.26&57.19&59.28&59.74&59.28&58.81&60.99&61.63&60.99&60.45\\
\cline{2-14}
&MonoLingual&62.66&65.15&62.66&60.97&60.97&62.06&60.97&60.01&64.82&65.63&64.82&64.31\\
\cline{2-14}
&MultiLingual&65.11&\textbf{66.33}&65.11&64.41&63.42&64.55&63.42&62.64&\textbf{65.14}&65.64&\textbf{65.14}&\textbf{64.83}\\
\hline
\hline
\multirow{3.5}{*}{\rotatebox{90}{\parbox{0.75cm}{\textsc{mWNC}}}}&ZeroShot&63.04&64.00&63.04&62.38&62.08&62.81&62.08&61.53&63.02&64.02&63.02&62.34\\
\cline{2-14}
&MonoLingual&64.82&65.95&64.82&64.17&63.15&63.71&63.15&62.75&66.59&66.92&66.59&66.42\\
\cline{2-14}
&MultiLingual&66.72&\textbf{67.24}&66.72&66.46&65.49&65.75&65.49&65.34&\textbf{66.96}&67.03&\textbf{66.96}&\textbf{66.92}\\
\hline
    \end{tabular}
\end{table*}

\begin{table}[!t]
\scriptsize
\centering
\caption{Detailed Language-wise Bias Detection Results for Multilingual Setup.}
\label{tab:clf-entire-wikibias-results}
\begin{tabular}{|c|l|rrrr|rrrr|rrrr|}
\hline
\multicolumn{2}{|c|}{}& \multicolumn{4}{c|}{MuRIL}                                                                                                                & \multicolumn{4}{c|}{InfoXLM}                                                                                   & \multicolumn{4}{c|}{mDeBERTa}                                                                                                             \\ \cline{3-14}
\multicolumn{2}{|c|}{} & \multicolumn{1}{l|}{Acc}            & \multicolumn{1}{l|}{P}              & \multicolumn{1}{l|}{R}              & \multicolumn{1}{l|}{F1} & \multicolumn{1}{l|}{Acc}   & \multicolumn{1}{l|}{P}     & \multicolumn{1}{l|}{R}     & \multicolumn{1}{l|}{F1} & \multicolumn{1}{l|}{Acc}            & \multicolumn{1}{l|}{P}              & \multicolumn{1}{l|}{R}              & \multicolumn{1}{l|}{F1} \\ \hline \hline
\multirow{9}{*}{\rotatebox{90}{\textsc{mWikiBias}}} & bn  & \multicolumn{1}{r|}{64.55}          & \multicolumn{1}{r|}{\textbf{65.65}} & \multicolumn{1}{r|}{64.55}          & 63.92                   & \multicolumn{1}{r|}{62.17} & \multicolumn{1}{r|}{63.38} & \multicolumn{1}{r|}{62.17} & 61.30                   & \multicolumn{1}{r|}{\textbf{64.62}} & \multicolumn{1}{r|}{65.15}          & \multicolumn{1}{r|}{\textbf{64.62}} & \textbf{64.31}          \\ \cline{2-14} 
                           & en  & \multicolumn{1}{r|}{73.69}          & \multicolumn{1}{r|}{\textbf{74.74}} & \multicolumn{1}{r|}{73.69}          & 73.40                   & \multicolumn{1}{r|}{72.89} & \multicolumn{1}{r|}{73.74} & \multicolumn{1}{r|}{72.89} & 72.65                   & \multicolumn{1}{r|}{\textbf{74.19}} & \multicolumn{1}{r|}{74.57}          & \multicolumn{1}{r|}{\textbf{74.19}} & \textbf{74.09}          \\ \cline{2-14} 
                           & gu  & \multicolumn{1}{r|}{63.77}          & \multicolumn{1}{r|}{\textbf{64.93}} & \multicolumn{1}{r|}{63.77}          & 63.06                   & \multicolumn{1}{r|}{62.03} & \multicolumn{1}{r|}{63.25} & \multicolumn{1}{r|}{62.03} & 61.13                   & \multicolumn{1}{r|}{\textbf{63.91}} & \multicolumn{1}{r|}{64.35}          & \multicolumn{1}{r|}{\textbf{63.91}} & \textbf{63.63}          \\ \cline{2-14} 
                           & hi  & \multicolumn{1}{r|}{\textbf{65.31}} & \multicolumn{1}{r|}{\textbf{66.37}} & \multicolumn{1}{r|}{\textbf{65.31}} & 64.73                   & \multicolumn{1}{r|}{63.54} & \multicolumn{1}{r|}{64.60} & \multicolumn{1}{r|}{63.54} & 62.86                   & \multicolumn{1}{r|}{65.01}          & \multicolumn{1}{r|}{65.34}          & \multicolumn{1}{r|}{65.01}          & \textbf{64.82}          \\ \cline{2-14} 
                           & kn  & \multicolumn{1}{r|}{\textbf{64.33}} & \multicolumn{1}{r|}{\textbf{65.63}} & \multicolumn{1}{r|}{\textbf{64.33}} & 63.57                   & \multicolumn{1}{r|}{62.48} & \multicolumn{1}{r|}{63.50} & \multicolumn{1}{r|}{62.48} & 61.76                   & \multicolumn{1}{r|}{63.96}          & \multicolumn{1}{r|}{64.49}          & \multicolumn{1}{r|}{63.96}          & \textbf{63.64}          \\ \cline{2-14} 
                           & mr  & \multicolumn{1}{r|}{\textbf{62.74}} & \multicolumn{1}{r|}{\textbf{63.98}} & \multicolumn{1}{r|}{\textbf{62.74}} & 61.89                   & \multicolumn{1}{r|}{61.47} & \multicolumn{1}{r|}{62.52} & \multicolumn{1}{r|}{61.47} & 60.65                   & \multicolumn{1}{r|}{62.26}          & \multicolumn{1}{r|}{62.71}          & \multicolumn{1}{r|}{62.26}          & \textbf{61.92}          \\ \cline{2-14} 
                           & ta  & \multicolumn{1}{r|}{63.05}          & \multicolumn{1}{r|}{64.47}          & \multicolumn{1}{r|}{63.05}          & 62.12                   & \multicolumn{1}{r|}{61.99} & \multicolumn{1}{r|}{63.23} & \multicolumn{1}{r|}{61.99} & 61.07                   & \multicolumn{1}{r|}{\textbf{63.80}} & \multicolumn{1}{r|}{\textbf{64.55}} & \multicolumn{1}{r|}{\textbf{63.80}} & \textbf{63.33}          \\ \cline{2-14} 
                           & te  & \multicolumn{1}{r|}{\textbf{63.46}} & \multicolumn{1}{r|}{\textbf{64.90}} & \multicolumn{1}{r|}{\textbf{63.46}} & 62.56                   & \multicolumn{1}{r|}{60.81} & \multicolumn{1}{r|}{62.17} & \multicolumn{1}{r|}{60.81} & 59.69                   & \multicolumn{1}{r|}{63.34}          & \multicolumn{1}{r|}{63.99}          & \multicolumn{1}{r|}{63.34}          & \textbf{62.91}          \\ \cline{2-14}&\multicolumn{13}{c|}{\vspace{-5pt}}\\
                           \cline{2-14}
                           & avg & \multicolumn{1}{r|}{65.11}          & \multicolumn{1}{r|}{\textbf{66.33}} & \multicolumn{1}{r|}{65.11}          & 64.41                   & \multicolumn{1}{r|}{63.42} & \multicolumn{1}{r|}{64.55} & \multicolumn{1}{r|}{63.42} & 62.64                   & \multicolumn{1}{r|}{\textbf{65.14}} & \multicolumn{1}{r|}{65.64}          & \multicolumn{1}{r|}{\textbf{65.14}} & \textbf{64.83}          \\ \hline\hline
\multirow{9}{*}{\rotatebox{90}{\textsc{mWNC}}}      & bn  & \multicolumn{1}{r|}{\textbf{66.75}} & \multicolumn{1}{r|}{\textbf{67.34}} & \multicolumn{1}{r|}{\textbf{66.75}} & \textbf{66.47}          & \multicolumn{1}{r|}{65.01} & \multicolumn{1}{r|}{65.32} & \multicolumn{1}{r|}{65.01} & 64.83                   & \multicolumn{1}{r|}{66.46}          & \multicolumn{1}{r|}{66.53}          & \multicolumn{1}{r|}{66.46}          & 66.42                   \\ \cline{2-14} 
                           & en  & \multicolumn{1}{r|}{71.08}          & \multicolumn{1}{r|}{71.43}          & \multicolumn{1}{r|}{71.08}          & 70.96                   & \multicolumn{1}{r|}{71.57} & \multicolumn{1}{r|}{71.66} & \multicolumn{1}{r|}{71.57} & 71.54                   & \multicolumn{1}{r|}{\textbf{72.92}} & \multicolumn{1}{r|}{\textbf{72.92}} & \multicolumn{1}{r|}{\textbf{72.92}} & \textbf{72.91}          \\ \cline{2-14} 
                           & gu  & \multicolumn{1}{r|}{66.00}          & \multicolumn{1}{r|}{\textbf{66.48}} & \multicolumn{1}{r|}{66.00}          & 65.76                   & \multicolumn{1}{r|}{64.33} & \multicolumn{1}{r|}{64.63} & \multicolumn{1}{r|}{64.33} & 64.15                   & \multicolumn{1}{r|}{\textbf{66.42}} & \multicolumn{1}{r|}{66.46}          & \multicolumn{1}{r|}{\textbf{66.42}} & \textbf{66.40}          \\ \cline{2-14} 
                           & hi  & \multicolumn{1}{r|}{67.13}          & \multicolumn{1}{r|}{\textbf{67.53}} & \multicolumn{1}{r|}{67.13}          & 66.95                   & \multicolumn{1}{r|}{66.28} & \multicolumn{1}{r|}{66.44} & \multicolumn{1}{r|}{66.28} & 66.20                   & \multicolumn{1}{r|}{\textbf{67.45}} & \multicolumn{1}{r|}{67.47}          & \multicolumn{1}{r|}{\textbf{67.45}} & \textbf{67.44}          \\ \cline{2-14} 
                           & kn  & \multicolumn{1}{r|}{66.40}          & \multicolumn{1}{r|}{\textbf{66.92}} & \multicolumn{1}{r|}{66.40}          & 66.14                   & \multicolumn{1}{r|}{64.77} & \multicolumn{1}{r|}{65.04} & \multicolumn{1}{r|}{64.77} & 64.61                   & \multicolumn{1}{r|}{\textbf{66.55}} & \multicolumn{1}{r|}{66.61}          & \multicolumn{1}{r|}{\textbf{66.55}} & \textbf{66.52}          \\ \cline{2-14} 
                           & mr  & \multicolumn{1}{r|}{\textbf{64.90}} & \multicolumn{1}{r|}{\textbf{65.48}} & \multicolumn{1}{r|}{\textbf{64.90}} & \textbf{64.57}          & \multicolumn{1}{r|}{63.70} & \multicolumn{1}{r|}{63.94} & \multicolumn{1}{r|}{63.70} & 63.54                   & \multicolumn{1}{r|}{64.44}          & \multicolumn{1}{r|}{64.56}          & \multicolumn{1}{r|}{64.44}          & 64.37                   \\ \cline{2-14} 
                           & ta  & \multicolumn{1}{r|}{\textbf{65.70}} & \multicolumn{1}{r|}{\textbf{66.29}} & \multicolumn{1}{r|}{\textbf{65.70}} & 65.38                   & \multicolumn{1}{r|}{64.36} & \multicolumn{1}{r|}{64.69} & \multicolumn{1}{r|}{64.36} & 64.15                   & \multicolumn{1}{r|}{65.68}          & \multicolumn{1}{r|}{65.83}          & \multicolumn{1}{r|}{65.68}          & \textbf{65.60}          \\ \cline{2-14} 
                           & te  & \multicolumn{1}{r|}{\textbf{65.78}} & \multicolumn{1}{r|}{\textbf{66.43}} & \multicolumn{1}{r|}{\textbf{65.78}} & 65.44                   & \multicolumn{1}{r|}{63.93} & \multicolumn{1}{r|}{64.30} & \multicolumn{1}{r|}{63.93} & 63.69                   & \multicolumn{1}{r|}{65.75}          & \multicolumn{1}{r|}{65.87}          & \multicolumn{1}{r|}{65.75}          & \textbf{65.68}          \\ \cline{2-14} 
                           &\multicolumn{13}{c|}{\vspace{-5pt}}\\
                           \cline{2-14}
                           & avg & \multicolumn{1}{r|}{66.72}          & \multicolumn{1}{r|}{\textbf{67.24}} & \multicolumn{1}{r|}{66.72}          & 66.46                   & \multicolumn{1}{r|}{65.49} & \multicolumn{1}{r|}{65.75} & \multicolumn{1}{r|}{65.49} & 65.34                   & \multicolumn{1}{r|}{\textbf{66.96}} & \multicolumn{1}{r|}{67.03}          & \multicolumn{1}{r|}{\textbf{66.96}} & \textbf{66.92}          \\ \hline
\end{tabular}
\end{table}

\section{Results}
\noindent\textbf{Bias Detection Results}: 
We show a summary of bias detection results, averaged across the 8 languages, in Table~\ref{tab:classifier-baselines-mWIKIBIAS-mWNC} and language-wise details in Table~\ref{tab:clf-entire-wikibias-results}. Table~\ref{tab:classifier-baselines-mWIKIBIAS-mWNC} shows that (1) Multilingual models outperform monolingual models, which in turn outperform zero-shot approaches. (2) Across both the datasets, mDeBERTa and MuRIL, both trained in a multilingual setting, exhibit the strongest performance, with mDeBERTa slightly outperforming MuRIL.

From the language-wise results in Table~\ref{tab:clf-entire-wikibias-results}, we observe the following: (1) As expected, for both datasets, across all models and metrics, best results are for English. We also observe that the models perform the worst for Marathi and Telugu. (2) In general, MuRIL is better in terms of precision, but mDeBERTa is better in terms of recall and also F1. 


\noindent\textbf{Bias Mitigation Results}: We show a summary of bias mitigation results in Table~\ref{tab:st-baselines-mWIKIBIAS-mWNC} and language-wise bias mitigation results in Table~\ref{tab:st-entire-results} using $N$=73.52 and 76.17 for \datasets{} respectively. From the results, we observe that (1) Broadly, multilingual models outperform monolingual counterparts. (2) mT5 is better for \textsc{mWikiBias} providing a high HM of 85.82, while mT0 is better for \textsc{mWNC} providing a high HM of 79.70. (3) As expected, both the models work best for English. 

\begin{table}[!t]
    \centering
    \scriptsize
    \caption{Bias Mitigation Results. B=BLEU, M=METEOR, C=chrF, BS=BERTScore, NAcc=NormAcc, HM=Harmonic Mean of BS and NAcc.}
\label{tab:st-baselines-mWIKIBIAS-mWNC}
    \begin{tabular}{|l|l|c|c|c|c|c|c|c|c|c|c|c|c|}
    \hline
\multicolumn{2}{|c|}{}&\multicolumn{6}{c|}{MonoLingual}&\multicolumn{6}{c|}{MultiLingual}\\
\cline{3-14}
\multicolumn{2}{|c|}{}&B&M&C&BS&NAcc&HM&B&M&C&BS&NAcc&HM\\
\hline
\hline
\multirow{3.5}{*}{\rotatebox{90}{\parbox{0.75cm}{\textsc{mWiki Bias}}}}&IndicBART&\textbf{63.67}&75.87&80.04&91.58&71.57&80.35&46.32&64.62&68.94&88.47&73.84&80.50\\
\cline{2-14}
&mT0&61.57&77.05&80.84&93.24&76.77&84.21&60.86&77.04&80.89&93.20&77.73&84.76\\
\cline{2-14}
&mT5&58.81&76.74&80.23&92.97&73.23&81.93&63.26&\textbf{77.41}&\textbf{81.39}&\textbf{93.40}&\textbf{79.38}&\textbf{85.82}\\
\hline
\hline
\multirow{3.5}{*}{\rotatebox{90}{\parbox{0.75cm}{\textsc{mWNC}}}}&IndicBART&54.98&69.25&75.27&90.99&65.52&76.18&17.58&59.67&61.15&85.54&\textbf{71.12}&77.67\\
\cline{2-14}
&mT0&53.09&70.01&75.75&91.27&69.15&78.68&55.23&\textbf{70.61}&\textbf{76.54}&\textbf{91.50}&70.59&\textbf{79.70}\\
\cline{2-14}
&mT5&\textbf{55.39}&70.28&76.22&91.36&66.57&77.02&55.27&70.46&76.41&91.47&69.83&79.20\\
\hline
    \end{tabular}
\end{table}

\begin{table*}[!t]
    \centering
    \scriptsize
    \caption{Detailed Language-wise Bias Mitigation Resultsfor the best models per dataset. B=BLEU, M=METEOR, C=chrF, BS=BERTScore, NAcc=NormAcc, HM=Harmonic Mean of BS and NAcc.}
\label{tab:st-entire-results}
    \begin{tabular}{|l|c|c|c|c|c|c|c|c|c|c|c|c|}
    \hline
&\multicolumn{6}{c|}{mT5 \textsc{mWikiBias}}&\multicolumn{6}{c|}{mT0 \textsc{mWNC}}\\
\cline{2-13}
&B&M&C&BS&NAcc&HM&B&M&C&BS&NAcc&HM\\
\hline
\hline
bn&60.60&75.82&80.03&92.81&76.50&83.87&54.76&68.48&75.10&90.72&70.29&79.21\\
\hline
en&86.02&92.68&91.63&98.30&88.06&92.90&79.06&87.56&87.38&97.42&79.97&87.83\\
\hline
gu&61.35&76.39&79.54&92.85&78.02&84.79&55.47&69.56&74.73&90.79&67.34&77.32\\
\hline
hi&69.36&82.87&82.76&94.16&75.78&83.97&63.12&76.81&77.85&92.19&69.49&79.25\\
\hline
kn&60.84&75.63&82.05&93.28&78.31&85.14&54.69&68.88&77.55&91.40&67.26&77.49\\
\hline
mr&58.19&73.25&78.11&92.12&79.43&85.31&50.24&65.44&72.65&89.77&68.78&77.89\\
\hline
ta&53.03&69.70&78.15&91.57&80.26&85.54&35.66&61.91&72.51&89.37&71.66&79.54\\
\hline
te&56.72&72.97&78.86&92.11&78.67&84.86&48.84&66.21&74.55&90.31&70.00&78.87\\
\hline
\hline
avg&63.26&77.41&81.39&93.40&79.38&85.82&55.23&70.61&76.54&91.50&70.59&79.70\\
\hline
    \end{tabular}
\end{table*}

\noindent\textbf{Human Evaluation Results}: 
We asked 4 Computer Science bachelors students with language expertise to evaluate the generated outputs (mT5 multilingual for \textsc{mWIKIBIAS} and mT0 multilingual for \textsc{mWNC}) on 3 criteria, each on a scale of 1 to 5: fluency, whether the bias is reduced and whether the meaning is preserved when compared to input. This was done for 50 samples per language for both datasets. Table~\ref{tab:humanEval} shows that automated evaluation correlates well with human judgment, with English predictions showing the best results. \textsc{mWNC} is easier for the models to debias than \textsc{mWIKIBIAS}.
The model outputs were generally fluent and had similar content as the input text. However, a wider variance in bias mitigation abilities was observed for the 3 Indian languages tested compared to English. Ambiguity in bias assessment and noise in the reference text made $\sim$20\% of the samples challenging for human annotators.

\begin{table}[!t]
    \centering
    \scriptsize
        \caption{Human Evaluation Results}
    \label{tab:humanEval}
    \begin{tabular}{|l|c|c|c|c|c|c|}
\hline
\multirow{2}{*}{Lang.}&\multicolumn{3}{c|}{\textsc{mWikiBias}}&\multicolumn{3}{c|}{\textsc{mWNC}}\\
\cline{2-7}
&Fluency ($\uparrow$)&Bias ($\downarrow$)&Meaning ($\uparrow$)&Fluency ($\uparrow$)&Bias ($\downarrow$)&Meaning ($\uparrow$)\\
\hline
\hline
bn&4.42&3.12&4.79&3.94&2.68&4.80\\
\hline
en&\textbf{4.92}&2.72&\textbf{4.84}&\textbf{4.86}&\textbf{2.40}&\textbf{4.92}\\
\hline
hi&4.20&3.20&4.76&4.60&2.64&\textbf{4.92}\\
\hline
te&4.40&\textbf{2.50}&4.81&3.88&2.45&4.75\\
\hline
\end{tabular}
\end{table}



\section{Conclusion}
In this paper, we proposed the critical problems of multilingual bias detection and mitigation for Indian languages. We also contributed two data sets, \datasets{}. We presented baseline results using standard Transformer based models. We make our code and data set publicly available\footref{datafootnote}. In the future, we would like to experiment with reinforcement learning based methods which could use detection based scores to enhance generation.

\bibliographystyle{splncs04}
\bibliography{references}

\end{document}